\documentclass[12pt]{l4dc2021}
\makeatletter
\def\set@curr@file#1{\def\@curr@file{#1}} 
\makeatother
\usepackage{hyperref}
\usepackage{graphicx}		
\usepackage{wrapfig}
\usepackage[format=plain,font=footnotesize,labelfont=bf,labelsep=period]{caption}
\usepackage[export]{adjustbox}

\usepackage{amsmath}
\usepackage{amssymb}
\usepackage{mathtools}
\usepackage{algorithm}
\usepackage{algpseudocode}

\usepackage{pdfsync}
\usepackage[normalem]{ulem}
\usepackage{paralist}	
\usepackage[space]{grffile} 

\usepackage{color}

\newcommand{\R}{\mathbb{R}}

\newcommand{\q}{q}

\newcommand{\C}{\mathcal{C}}

\definecolor{darkblue}{RGB}{0,0,102}
\definecolor{lightblue}{RGB}{77,77,148}

\definecolor{gold}{RGB}{234, 170, 0}
\definecolor{metallic_gold}{RGB}{139, 111, 78}

\renewcommand{\cal}[1]{\mathcal{ #1 }}
\newcommand{\mb}[1]{\mathbf{ #1 }}
\newcommand{\bs}[1]{\boldsymbol{ #1 }}

\newcommand{\grad}{\nabla}

\DeclareMathOperator*{\argmin}{argmin}

\DeclareMathOperator*{\esssup}{ess\,sup}

\newcommand\blfootnote[1]{%
  \begingroup
  \renewcommand\thefootnote{}\footnote{#1}%
  \addtocounter{footnote}{-1}%
  \endgroup
}

\title[Learning Safe Bipedal Locomotion]{
Episodic Learning for Safe Bipedal Locomotion with \\ Control Barrier Functions and Projection-to-State Safety
}
\usepackage{times}
\usepackage[export]{adjustbox}
\newcommand{\f}{\mb{f}}
\newcommand{\g}{\mb{g}}
\newcommand{\x}{\mb{x}}
\renewcommand{\u}{\mb{u}}

\newcommand{\y}{\mb{y}}

\renewcommand{\q}{\mb{q}}
\newcommand{\K}{\mathcal{K}}
\author{%
\Name{Noel Csomay-Shanklin$^*$}
\Email{noelcs@caltech.edu} \\
\Name{Ryan K. Cosner$^*$}
\Email{rkcosner@caltech.edu} \\
\Name{Min Dai$^*$}
\Email{mdai@caltech.edu} \\
\Name{Andrew J. Taylor}
\Email{ajtaylor@caltech.edu} \\
\Name{Aaron D. Ames}
\Email{ames@caltech.edu} \\
}
\begin{document}
\maketitle

\vspace{-10pt}
\begin{abstract} 
This paper combines episodic learning and control barrier functions in the setting of bipedal locomotion. The safety guarantees that control barrier functions provide are only valid with perfect model knowledge; however, this assumption cannot be met on hardware platforms. To address this, we utilize the notion of projection-to-state safety paired with a machine learning framework in an attempt to learn the model uncertainty as it affects the barrier functions. The proposed approach is demonstrated both in simulation and on hardware for the AMBER-3M bipedal robot in the context of the stepping-stone problem, which requires precise foot placement while walking dynamically.
%
%
%
\end{abstract}
%
\blfootnote{All authors are with Caltech, Pasadena, CA, USA. $^*$ These authors contributed equally to this work.}
\section{Introduction} 
\label{sec:intro}

Safety is of significant importance in many modern control applications, with complex systems demanding the rigorous encoding of safety properties in the controller design process. One particularly challenging example is the safe locomotion of bipedal robots in difficult terrain; a concrete example of this is the ``stepping-stone problem'' where the feet must be placed in precise locations. In practice, the models used to design safety-critical controllers are imperfect---especially in the context of highly dynamic systems like walking robots---with model uncertainty arising due to parametric error and unmodeled dynamics. This uncertainty can lead to unsafe or dangerous behavior, thus requiring that safety-critical control be understood in the presence of model uncertainty.

In this work we consider a machine learning based approach for attenuating the impact of model uncertainty on the safe behavior of a system. The application of learning to safety-critical control of systems with uncertain models has recently exhibited significant success \citep{taylor2019learning2,cheng2019accelerating, choi2020reinforcement, lee2020learning, castaneda2020gaussian}. We look to achieve safety defined in terms of \textit{set invariance} \citep{blanchini1999set}, which is an area of active research at the intersection of data-driven methods and nonlinear control theory \citep{berkenkamp2016safe2, fisac2018general, wang2018safe, fan2019bayesian, dean2020guaranteeing}. 

We will leverage control barrier functions (CBFs) for synthesizing safety controllers \citep{ames2014control, ames2017control} that achieve set invariance. CBFs have become a popular method for ensuring safety \citep{nguyen2016exponential, wang2018safe}, but they require an accurate model of the system dynamics and model uncertainty can result in a loss of safety guarantees \citep{kolathaya2018input}. Efforts have been made to make the safety guarantees of CBFs robust to the effects of model uncertainty, but in doing so these methods can cause the system's behavior to be overly conservative \citep{gurriet2018towards, xu2015robustness, taylor2019adaptive}. 

One challenge in incorporating learning methods with nonlinear control is the need for diverse data capturing system behavior, and in particular the input-to-state relationship. While in simulated environments it is possible to collect high coverage data sets that accurately determine how inputs affect the evolution of the system, collecting such data on real physical systems may be prohibitively costly or damaging to the system. The lack of this data can lead to challenges with under-determination in supervised learning problems that seek to preserve the underlying structure of dynamic systems \citep{taylor2019episodic, taylor2019learning2}. Often this leads to model fits with low training loss that result in poor control performance when integrated in closed-loop controllers.

One setting for which collecting diverse data is infeasible is bipedal locomotion, as consistent walking requires well defined input profiles to ensure stability \citep{grizzle2001asymptotically}. The stepping-stone task is a historical benchmark for evaluating the safety-critical control of biped platforms \citep{nguyen20163d, nguyen2018dynamic, nguyen2020dynamic}. As compared to the statically stable motion which has been successfully employed to accomplish this task on a variety of legged robotic platforms \citep{griffin2019footstep, fankhauser2018robust}, the underactuated nature of the bipedal robot under consideration in this work requires dynamic gaits to successfully traverse a set of stepping-stones. During such dynamic motion, satisfaction of safety constraints is predicated on having accurate model information, a requirement that cannot be met on hardware platforms, thus suggesting the need for model uncertainty reduction via learning.

\textbf{Our Contribution:} The main contributions of this work are two-fold. First, we propose an episodic learning framework for iteratively reducing the impact of disturbances on the safety-critical behavior of a system. Compared to previous work, our method resolves challenges with input data diversity by recursively defining disturbance estimators and controllers that are agnostic to the underlying control-affine structure. We successfully deploy this method in the context of the classical bipedal robotic stepping-stone problem. We validate that this method is able to effectively reduce model uncertainty and improve the safe behavior of a complex system both in simulation and experiment. The second contribution is the first experimental demonstration of CBFs for safety-critical control on a bipedal robot. This challenge has been the focus of significant prior work and we demonstrate that it can be achieved through the introduction of learning.




\section{Background} 
\label{sec:background}
In this section we provide a review of CBFs and projection-to-state safety (PSSf), which will be utilized in Section \ref{sec:learning} to formulate a learning approach for mitigating model uncertainty.

\subsection{Safety via Control Barrier Functions}
Consider the nonlinear control affine system given by:
\begin{equation}
    \label{eqn:eom}
    \dot{\mb{x}} = \mb{f}(\mb{x})+\mb{g}(\mb{x})\mb{u},
\end{equation}
where $\mb{x}\in\R^n$, $\mb{u}\in\R^m$, and $\mb{f}:\R^n\to\R^n$ and $\mb{g}:\R^n\to\R^{n\times m}$ are locally Lipschitz continuous on $\R^n$. Given a Lipschitz continuous state-feedback controller $\mb{k}:\R^n\to\R^m$, the closed-loop system dynamics are:
\begin{equation}
    \label{eqn:cloop}
    \dot{\mb{x}} = \mb{f}_{\textrm{cl}}(\mb{x}) \triangleq  \mb{f}(\mb{x})+\mb{g}(\mb{x})\mb{k}(\mb{x}).
\end{equation}
The assumption of local Lipschitz continuity of $\mb{f}$, $\mb{g}$, and $\mb{k}$ implies that $\mb{f}_\textrm{cl}$ is locally Lipschitz continuous. Thus, for any initial condition $\mb{x}_0 := \mb{x}(0) \in \R^n$ there exists a maximum time interval $I(\mb{x}_0) = [0, t_{\textrm{max}})$ such that $\mb{x}(t)$ is the unique solution to \eqref{eqn:cloop} on $I(\mb{x}_0)$ \citep{perko2013differential}. In the case that $\mb{f}_{\textrm{cl}}$ is forward complete, $t_{\textrm{max}}=\infty$.

The notion of safety that we consider in this paper is formalized by specifying a \textit{safe set} in the state space that the system must remain in to be considered safe. In particular, consider a set $\C\subset \R^n$ defined as the 0-superlevel set of a continuously differentiable function $h:\R^n\to \R$, yielding:
\begin{align}
    \C &\triangleq \left\{\mb{x} \in \R^n : h(\mb{x}) \geq 0\right\}, \label{eqn:safeset}
\end{align}
with $\partial\C \triangleq \{\mb{x} \in \R^n : h(\mb{x}) = 0\}$ and
    $\textrm{Int}(\C) \triangleq \{\mb{x} \in \R^n : h(\mb{x}) > 0\}$. We assume that $\C$ is nonempty and has no isolated points, that is, $\textrm{Int}(\C) \not = \emptyset \textrm{ and }\overline{\textrm{Int}(\C)} = \C$. We refer to $\C$ as the \textit{safe set}. This construction motivates the following definitions of forward invariance and safety:

\begin{definition}[\textit{Forward Invariance \& Safety}]
A set $\C\subset\R^n$ is \textit{forward invariant} if for every $\mb{x}_0\in\C$, the solution $\mb{x}(t)$ to \eqref{eqn:cloop} satisfies $\mb{x}(t) \in \C$ for all $t \in I(\mb{x}_0)$. The system \eqref{eqn:cloop} is \textit{safe} on the set $\C$ if the set $\C$ is forward invariant.
\end{definition}

Before defining CBFs, we denote a continuous function $\alpha:[0,a)\to\R_+$, with $a>0$, as \textit{class $\cal{K}$} ($\alpha\in\cal{K}$) if $\alpha(0)=0$ and $\alpha$ is strictly monotonically increasing. If $a=\infty$ and $\lim_{r\to\infty}\alpha(r)=\infty$, then $\alpha$ is \textit{class $\cal{K}_\infty$} ($\alpha\in\cal{K}_\infty$). A continuous function $\alpha:(-b,a)\to\R$, with $a,b>0$, is said to belong to \textit{extended class $\cal{K}$} ($\alpha\in\cal{K}_e$) if $\alpha(0)=0$ and $\alpha$ is strictly monotonically increasing. If $a,b=\infty$, $\lim_{r\to\infty}\alpha(r)=\infty$, and $\lim_{r\to-\infty}\alpha(r)=-\infty$, then $\alpha$ is said to belong to \textit{extended class $\cal{K}_\infty$} ($\alpha\in\cal{K}_{\infty,e}$).

Certifying the safety of the closed-loop system \eqref{eqn:cloop} with respect to a set $\C$ may be impossible if the controller $\mb{k}$ was not chosen to enforce the safety of $\C$. CBFs can serve as a synthesis tool for attaining the forward invariance, and thus the safety of a set:
\begin{definition}[\textit{Control Barrier Function (CBF)} \citep{ames2017control}]
Let $\C\subset\R^n$ be the 0-superlevel set of a continuously differentiable function $h:\R^n\to\R$ with $0$ a regular value. The function $h$ is a \textit{control barrier function} (CBF) for \eqref{eqn:eom} on $\C$ if there exists $\alpha\in\K_{\infty,e}$ such that for all $\mb{x}\in\R^n$:
\begin{align}
\label{eqn:cbf}
     \sup_{\mb{u}\in\R^m} \dot{h}(\mb{x},\mb{u}) \triangleq \grad h(\mb{x})\left(\mb{f}(\mb{x})+\mb{g}(\mb{x})\mb{u}\right) = L_{\f}h(\x) + L_{\g}h(\x) \u \geq-\alpha(h(\mb{x})),
\end{align}
where $L_{\f}h(\x)$ and $L_{\g}h(\x)$ are Lie derivatives.
\end{definition}
\noindent Given a CBF $h$ for \eqref{eqn:eom} and a corresponding $\alpha\in\cal{K}_{\infty,e}$, we define the point-wise set of all control values that satisfy \eqref{eqn:cbf}:
\begin{equation*}
    K_{\textrm{cbf}}(\mb{x}) \triangleq \left\{\mb{u}\in\R^m ~\left|~ \dot{h}(\mb{x},\mb{u})\geq-\alpha(h(\mb{x})) \right. \right\}.
\end{equation*}
We have the following result relating controllers taking values in  $K_{\textrm{cbf}}(\mb{x})$ to the safety of \eqref{eqn:eom} on $\C$:
\begin{theorem}[\cite{ames2014control}]
Given a set $\C\subset\R^n$ defined as the 0-superlevel set of a continuously differentiable function $h:\R^n\to\R$, if $h$ is a CBF for \eqref{eqn:eom} on $\C$, then any Lipschitz continuous controller $\mb{k}:\R^n\to\R^m$, such that $\mb{k}(\mb{x})\in K_{\textrm{cbf}}(\mb{x})$ for all $\mb{x}\in\R^n$, renders the system \eqref{eqn:eom} safe with respect to the set $\C$.
\end{theorem}
Given a nominal (but not necessarily safe) locally Lipschitz continuous controller $\mb{k}_d:\R^n\to\R^m$, a controller taking values in the set $K_\textrm{cbf}(\mb{x})$ is the safety-critical CBF-QP:
\begin{align}
\label{eqn:CBF-QP}
\tag{CBF-QP}
\mb{k}(\mb{x}) =  \,\,\underset{\mb{u} \in \R^m}{\argmin}  &  \quad \frac{1}{2} \| \mb{u} -\mb{k}_d(\mb{x})\|_2^2  \\
\mathrm{s.t.} \quad & \quad \dot{h}(\mb{x},\mb{u})
\geq -\alpha(h(\mb{x})). \nonumber
\end{align}

\subsection{Model Uncertainty and Projection-to-State Safety}
\label{sec:uncertainty}

In practice, the system dynamics \eqref{eqn:eom} are not known during  control design  due to parametric error and unmodeled dynamics. Instead, a nominal model of the system is utilized:
\begin{equation}
\label{eqn:eomnominal}
    \widehat{\dot{\mb{x}}} = \widehat{\mb{f}}(\mb{x}) + \widehat{\mb{g}}(\mb{x})\mb{u},
\end{equation}
where $\widehat{\mb{f}}:\R^n\to\R^n$ and $\widehat{\mb{g}}:\R^n\to\R^{n\times m}$ are assumed to be Lipschitz continuous on $\R^n$. By adding and subtracting the right hand side of \eqref{eqn:eomnominal} to \eqref{eqn:eom}, the dynamics of the system are:
\begin{equation}
    \dot{\mb{x}} = \widehat{\mb{f}}(\mb{x}) + \widehat{\mb{g}}(\mb{x})\mb{u} + \overbrace{\underbrace{\mb{f}(\mb{x})-\widehat{\mb{f}}(\mb{x})}_{\mb{b}(\mb{x})}+\underbrace{\left(\mb{g}(\mb{x})-\widehat{\mb{g}}(\mb{x})\right)}_{\mb{A}(\mb{x})}\mb{u}}^{\mb{d}(\x,\u)},
\end{equation}
where the unknown disturbance $\mb{d}(\x,\u)=\mb{b}(\mb{x})+\mb{A}(\mb{x})\mb{u}$ is assumed to be time invariant, but depends on the state and input to the system. If the function $h:\R^n\to\R$ is a CBF for the nominal model \eqref{eqn:eomnominal} on $\C$, the uncertainty in the dynamics directly manifests in the time derivative of $h$:
\begin{align}
\label{eqn:hdotdecomp}
    \dot{h}(\mb{x},\mb{u})  =  \underbrace{\grad h(\mb{x})(\widehat{\mb{f}}(\mb{x})+\widehat{\mb{g}}(\mb{x})\mb{u})}_{\widehat{\dot{h}}(\mb{x},\mb{u})} 
    + \underbrace{\grad h(\mb{x})\mb{b}(\mb{x})}_{b(\mb{x})} + \underbrace{\grad h(\mb{x})\mb{A}(\mb{x})}_{\mb{a}(\mb{x})^\top}\mb{u}.
\end{align}
Given that $h$ is a CBF for \eqref{eqn:eomnominal} on $\C$, let $\mb{k}:\R^n\to\R^m$ be a Lipschitz continuous state-feedback controller such that $\widehat{\dot{h}}(\mb{x},\mb{k}(\mb{x})) \geq -\alpha(h(\mb{x}))$. Defining the \textit{projected disturbance} as: 
\begin{equation}
\label{eqn:projdist}
\delta(\mb{x}) \triangleq \dot{h}(\mb{x},\mb{k}(\mb{x})) - \widehat{\dot{h}}(\mb{x},\mb{k}(\mb{x})) = b(\mb{x})+\mb{a}(\mb{x})^\top\mb{k}(\mb{x}),
\end{equation}
yields:
\begin{equation}
\label{eqn:hdottrueineq}
    \dot{h}(\mb{x},\mb{k}(\mb{x})) \geq -\alpha(h(\mb{x})) - \delta(\mb{x})
\end{equation}
Assuming that $\delta$ is essentially bounded in time (there exists $M\in\R$, $M>0$, such that $\vert\delta\vert_\infty  \triangleq\esssup_{t\geq 0}\vert\delta(\mb{x}(t))\vert< M$), we may make use of the following definition:
\begin{definition}[\textit{Projection-to-State Safety (PSSf)} (\cite{taylor2020control})]
Given a feedback controller $\mb{k}$, the closed-loop system \eqref{eqn:cloop}, $\dot{\mb{x}} = \mb{f}_{\textrm{cl}}(\mb{x}) = \widehat{\mb{f}}(\mb{x}) + \widehat{\mb{g}}(\mb{x})\mb{k}(\mb{x}) + \mb{d}(\mb{x})$ with $\mb{d}(\mb{x})=\mb{b}(\mb{x})+\mb{A}(\mb{x})\mb{k}(\mb{x})$,  is \emph{projection-to-state safe} (PSSf) on $\C$ with respect to the function $h$ and projected disturbances $\delta:\R^n\to\R$ if there exists $\overline{\delta}>0$ and $\gamma\in\K_\infty$ such that the set $\C_{\delta}\supset\C$,
\begin{align}\label{eq:delta_safe_set}
    \C_{\delta} &\triangleq \left\{\mb{x} \in \R^n : h(\mb{x})+\gamma(\vert\delta\vert_\infty) \geq 0\right\},
\end{align}
is forward invariant for all $\delta$ satisfying $\vert\delta\vert_\infty\leq\overline{\delta}$.
\end{definition}

PSSf captures the fact that in the presence of model uncertainty, satisfying the CBF condition \eqref{eqn:cbf} for the estimated time derivative $\widehat{\dot{h}}$ is not sufficient for safety, as the projected disturbance $\delta$ appears in the lower bound on true time derivative of $h$ as in \eqref{eqn:hdottrueineq}. This results in a larger forward invariant set, given by $\mathcal{C}_\delta$, that grows with the magnitude of the projected disturbance.


\begin{figure}
    \centering
    \includegraphics{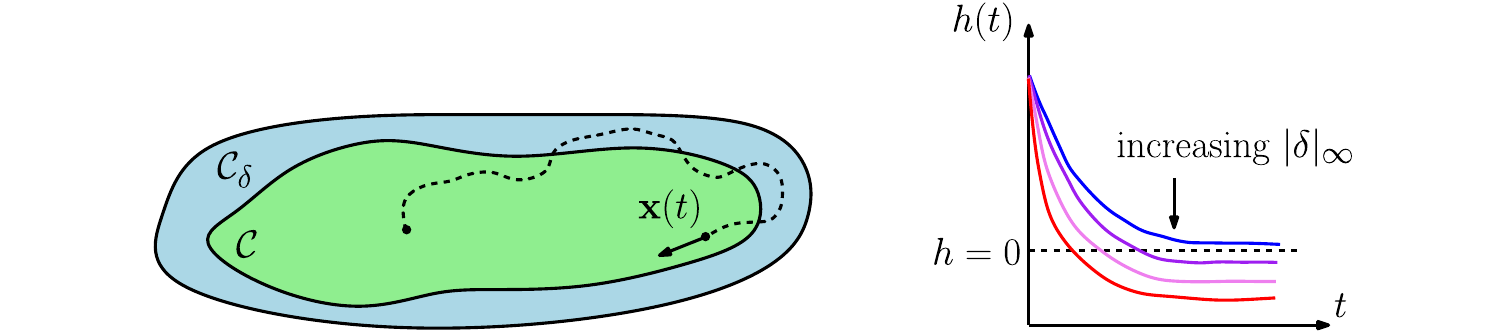}
    \caption{Geometric visualization of Projection-to-State Safety. The system is able to leave the safe set $\mathcal{C}$, but remains within the larger set $\mathcal{C}_\delta$. Maximum possible deviation from the safe set grows larger with $\vert\delta\vert_\infty$.}
    \label{fig:pssf}
    \vspace{-3mm}
\end{figure}



\section{Learning Projected Disturbances}
\label{sec:learning}
In this section we explore how an estimate of the projected disturbance $\delta$ can be learned episodically from data and incorporated into control synthesis to improve PSSf behavior. 

As seen in Section \ref{sec:background}, the projected disturbance $\delta$ appears in the time derivative of the barrier function $\dot{h}$, and potentially leads to unsafe behavior since it compromises the CBF condition as in \eqref{eqn:hdottrueineq}. If an upper bound $\overline{\delta}$ on $\vert\delta\vert_\infty$ is known (or determined heuristically), it could be directly incorporated into the inequality enforced in the controller:
\begin{align}
    \label{eqn:dumbreject}
    \tag{$\overline{\delta}$-CBF-QP}
    \mb{k}(\mb{x}) =  \,\,\underset{\mb{u} \in \R^m}{\argmin}  &  \quad \frac{1}{2} \| \mb{u}-\mb{k}_d(\mb{x}) \|_2^2  \\
    \mathrm{s.t.} \quad & \quad \widehat{\dot{h}}(\x, \u) - \overline{\delta}
    \geq -\alpha(h(\mb{x})). \nonumber
\end{align}
While this will enforce safety of the original set $\mathcal{C}$, it can be exceedingly conservative if $\overline{\delta}$ is larger than the actual projected disturbance. Furthermore, as the projected disturbance is a function of the state, its magnitude (and possibly sign) may change along a trajectory, leading to additional conservativeness in this approach.

Instead, we consider a learning approach to resolve the impact of $\delta$. To motivate such an approach, consider the following setting: in an experiment, the system is allowed to evolve forward in time from a particular initial condition and under a given state-feedback controller. During this experiment, data is collected which provides a discrete-time history of the CBF, $h$. This time history is smoothed and numerically differentiated to compute an approximate time history of the true value of the time derivative of the CBF, $\dot{h}$. This yields a collection of input-output pairs: 
\begin{equation}
    {D}_i = ((\mb{x}_i, \mb{k}(\x_i)), \dot{h}_i) \in (\R^n \times \R^m) \times \R
\end{equation}
whereby a dataset $\mathfrak{D}=\{{D}_i\}_{i=1}^N$ can be constructed. Given a nonlinear function class $\mathcal{H}: \R^n \to \R$ and a loss function $\mathcal{L}: \R \times \R \to \R $, a learning problem can be specified as finding a function $\widehat{\delta} \in \mathcal{H}$ to estimate $\delta$ via empirical risk minimization: 
\begin{equation}
    \label{eqn:erm}
    \tag{ERM}
    \inf_{\widehat{\delta} \in \mathcal{H}}\frac{1}{N}\sum_{i=1}^N\cal{L}\left(\widehat{\dot{h}}(\mb{x}_i,\mb{k}(\x_i))+ \widehat{\delta}(\x_i),\dot{h}_i\right).
\end{equation}
A controller can be synthesized which incorporates $\widehat{\delta}$ as follows: 
\begin{align}
    \label{eqn:delta-CBF-QP}
    \tag{$\widehat{\delta}$-CBF-QP}
    \mb{k}(\mb{x}) =  \,\,\underset{\mb{u} \in \R^m}{\argmin}  &  \quad \frac{1}{2} \| \mb{u}-\mb{k}_d(\mb{x}) \|_2^2  \\
    \mathrm{s.t.} \quad & \quad \widehat{\dot{h}}(\x, \u) + \widehat{\delta}(\x) 
    \geq -\alpha(h(\mb{x})). \nonumber
\end{align}
Note that compared with \ref{eqn:CBF-QP}, the extended safe set with \ref{eqn:delta-CBF-QP} shrinks from \eqref{eq:delta_safe_set} to  $\C_{\delta} = \left\{\mb{x} \in \R^n : h(\mb{x})+\gamma(\vert\delta-\hat{\delta}\vert_\infty) \geq 0\right\}$. 

We directly build upon the episodic learning framework from \citep{taylor2019episodic, taylor2019learning2} by seeking to learn $\delta$. Our approach is outlined in Algorithm 1.  In each episode, the  algorithm runs the current controller to collect data, learns a new $\widehat{\delta}$ using the newly collected data, and synthesizes a new controller. In this prior work, which was applied to less complex dynamical systems, the collected data was rich enough to determine a control affine structure. In many contexts, such as bipedal robots, such a degree of diversity is infeasible without damaging the system. We instead directly learn $\delta$ as a function of the previous controller $\mb{k}$ via a recursive relationship, as updating the estimator leads to the definition of a new projected disturbance $\delta' = b(\mb{x})+\mb{a}(\mb{x})^\top\mb{k}'(\mb{x})$. This yields a projected disturbance $\delta$  learned iteratively by modifying $\widehat{\delta}$ over the course of multiple episodes. This episodic approach to safety-critical control is captured in Algorithm \ref{alg:episodic_learning}.

\begin{algorithm}[H]
    \SetAlgoLined
    \caption{Projected Disturbance Learning (PDL)}
    \SetKwFunction{Sample}{sample}\SetKwFunction{Experiment}{experiment} \SetKwFunction{ERM}{ERM}\SetKwFunction{Augment}{augment}
    \SetKwFunction{LCBFQP}{$\widehat{\delta}$-CBF-QP}
    \SetKwInOut{Input}{input} \SetKwInOut{Output}{output}
    \label{alg:episodic_learning}
    \Input{CBF $h$, CBF derivative estimate $\widehat{\dot{h}}$, model class $\mathcal{H}$, loss function $\mathcal{L}$, nominal state-feedback controller $\mb{k}_0$, number of episodes $T$, initial condition $\x_0$}
    \Output{Augmented Controller $\mb{k}_T$}
    \BlankLine
    \For{$j = 1, \ldots, T $}{
        $\mathfrak{D}_j \leftarrow$\Experiment{$\mb{x}_0,\mb{k}_{j-1}$}  \hspace{2.438cm}\tcp*[h]{Execute experiment}\;\\
        $\widehat{\delta}\leftarrow$\ERM{$\cal{H}, \cal{L},\mathfrak{D}_j,\widehat{\dot{h}}_0$}  \hspace{3.7cm}\tcp*[h]{Fit estimator}\;\\
        $\widehat{\dot{h}}_j\leftarrow \widehat{\dot{h}}_0+\widehat{\delta}$  \hspace{5.49cm}\tcp*[h]{Update derivative estimator}\;\\
        $\mb{k}_j\leftarrow$ \LCBFQP{$\widehat{\dot{h}}_j$} \hspace{3.9cm}\tcp*[h]{Synthesize new controller}\;
    }
\end{algorithm}

\section{Bipedal Robotics} 
\label{sec:bipeds}
In this section we specify the notion of learning projected disturbances to the setting of bipedal locomotion. We briefly introduce the theory of bipedal locomotion and then describe the barrier function formulations which allow us to achieve safe bipedal locomotion across stepping-stones. A deeper exploration of this material may be found in \citep{westervelt2003hybrid}.

\subsection{Bipedal Robotic Dynamics}
\label{sec:bipedsdynamics}
\begin{figure}[ht!]
\centering
\includegraphics[width=\textwidth]{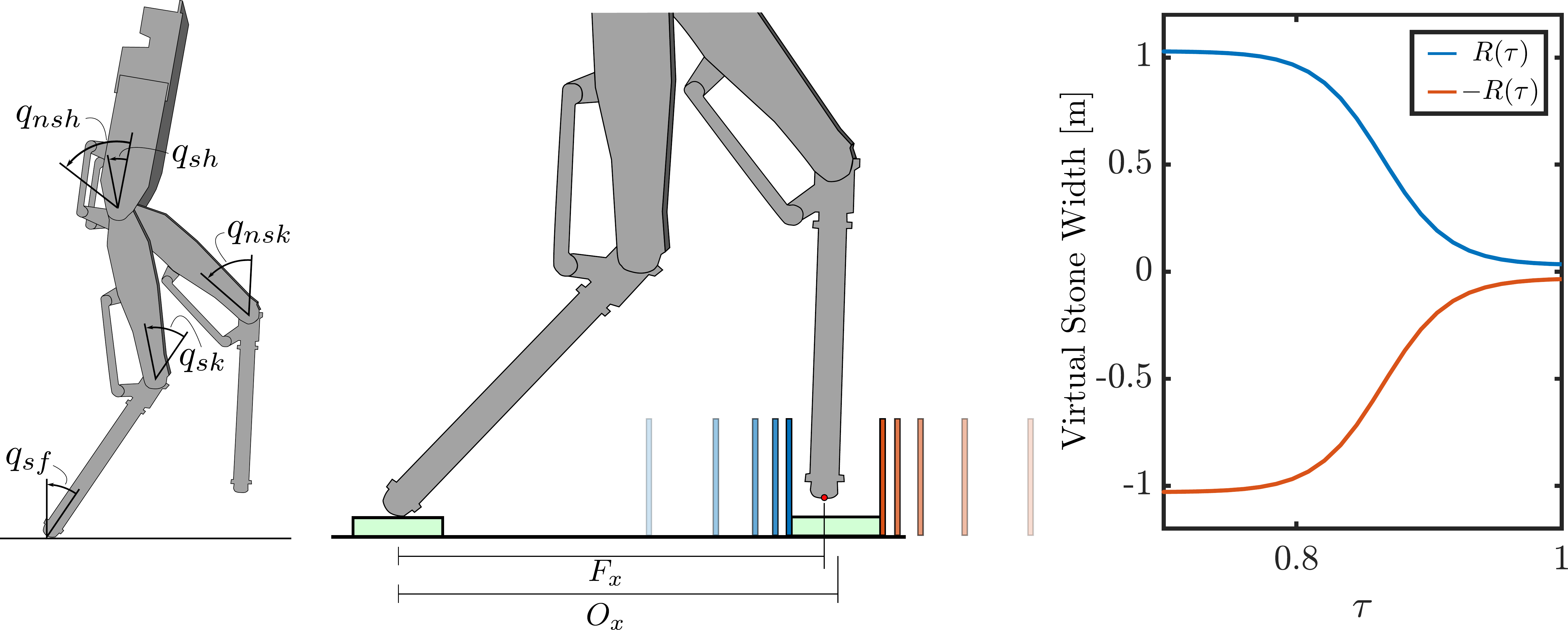}
	\caption{\textbf{(Left):} Schematic diagram of the AMBER-3M robot with position coordinates. \textbf{(Center):} Schematic of the foot placement in the stepping-stone problem. The boundaries of virtual stepping-stones are captured via the blue and orange vertical lines. \textbf{(Right):} Virtual stepping stone width as function of the phase variable $\tau(\mb{q})$.}
	\label{fig:amber_and_barrier}
\end{figure}

The bipedal robotic system we consider is the AMBER-3M robotic platform seen in Figure \ref{fig:amber_and_barrier}, modeled as an underactuated, planar five-link robot with point feet \citep{ambrose2017toward} whose physical parameters are reported in \cite[Table 1]{ma2014human}. The configuration coordinates $\mb{q}\in \mathcal{Q}\subset \R^5$ are given by
$
    \mb{q} = [q_{sf},\ q_{sk},\ q_{sh},\ q_{nsh},\ q_{nsk}]^\top,
$
with stance foot angle $q_{sf}$, stance knee angle $q_{sk}$, stance hip angle $q_{sh}$, non-stance hip angle $q_{nsh}$ and non-stance knee angle $q_{nsk}$. The continuous-time equations of motion, derived from the Euler-Lagrange equations, are given by: 
\begin{align}
    \mb{D}(\mb{q})\ddot{\mb{q}}+\mb{C}(\mb{q},\dot{\mb{q}})\dot{\mb{q}}+\mb{G}(\mb{q})&=\mb{Bu}, \label{roboDyn}
\end{align}
where $\mb{D}(\q)\in{\mathbb{S}}_{++}^{5}$ is the positive definite mass-inertia matrix,  
$\mb{C}(\q,\dot{\q}) \in \R^{5\times 5}$ contains the centrifugal and Coriolis forces, $\mb{G}(\q)\in \R^{5}$ contains the gravitational forces, $\mb{B}\in \R^{5\times 4}$ is the actuation matrix, and $\mb{u}\in\mathcal{U}\subset\mathbb{R}^4$ is the input. Note that this is the pinned model of the robot dynamics; for the unpinned model, refer to \citep{hereid2018dynamic}. For AMBER-3M, the number of inputs is one fewer than the degrees of freedom, meaning the system has one degree of underactuation.

Taking $p^v:\mathcal{Q}\to \mathbb{R}$ to represent the vertical position (height) of the swing foot, the admissible states are given by the \textit{domain} $\mathcal{D} = \{(\q,\mb{\dot{q}})\in T\mathcal{Q}\ |\ p^v(\mb{q}) \ge 0\}$. The switching surface on which the impact events occur, also known as the \textit{guard}, is defined by:
\begin{align}
    \mathcal{S} = \{(\mb{q},\dot{\mb{q}})\in T\mathcal{Q} ~|~ p^v(\q) = 0, \dot{p}^v(\q,\dot{\q}) < 0\}\subset \mathcal{D}.
\end{align}
With the full system state given by $\mb{x}=(\mb{q},\dot{\mb{q}})\in T\mathcal{Q}$, the impact dynamics \citep{westervelt2007feedback} are defined by a \textit{reset map} $\bs{\Delta}:\mathcal{S}\to \mathcal{D}$ relating pre-impact states $\mb{x}^-(t) \triangleq \lim_{\tau \nearrow t} \mb{x}(\tau)$ and post-impact states  $\mb{x}^+(t) = \lim_{\tau \searrow t} \mb{x}(\tau)$ via $\mb{x}^+(t) = \bs{\Delta}(\mb{x}^-(t))$. Combining these concepts and rearranging (\ref{roboDyn}) into control affine form yields the following \textit{hybrid control system}:
\begin{equation}\label{eq:hybrid_model_1}
\mathcal{HC} = \begin{cases} \dot{\mb{x}} = \mb{f}(\mb{x}) + \mb{g}(\mb{x}) \mb{u} \quad &\mb{x}^- \in \mathcal{D} \setminus \mathcal{S},\\  
\mb{x}^+= \bs{\Delta}(\mb{x}^-) &\mb{x}^- \in \mathcal{S}.\end{cases}    
\end{equation}
\subsection{Bipedal Robotic Control}
\label{sec:bipedcontrol}
Control of bipedal robotic systems centers around a \textit{phasing variable}, $\tau:\mathcal{Q}\to [0,1]$, given by:
\begin{equation} \label{eq:phase_variable}
    \tau(\mb{q}) = \frac{\delta_{hip}(\mb{q}) - \delta_{hip}^+}{\delta_{hip}^- -\delta_{hip}^+},
\end{equation}
where $\delta_{hip}:\R^5\to\R$ defined as $\delta_{hip}(\mb{q}) = [-l_t-l_f,\ -l_f,\ 0,\ 0,\ 0]\mb{q}$ is the linearized hip position with $l_t$ and $l_f$ the length of the robots tibia and femur, respectively. The constants \(\delta_{hip}^+\) and \(\delta_{hip}^-\) are the linearized hip positions at the beginning and the end of a step, ensuring that $\tau(\mb{q})$ increases monotonically in time within a step. Desired trajectories resulting in walking gaits for the robot can be rapidly synthesized via a hybrid zero dynamics framework (\cite{westervelt2003hybrid,ames2014rapidly}).
We are now well equipped to define the relative degree 2 (\cite{sastry1999nonlinear}) outputs $\mb{y}:\mathcal{Q}\to \mathbb{R}^4$ as the difference between the actual output $\mb{y}_a$ and the desired output trajectory $\mb{y}_d$:
\begin{equation}
    \mb{y}(\mb{q},\bs{\alpha}) \triangleq \mb{y}_a(\mb{q}) - \mb{y}_d(\tau(\mb{q}),\bs{\alpha}),
\end{equation}
with $\bs{\alpha}$ being the coefficients of a Bézier polynomial coming from the trajectory generation step. The actual output is given by the actuated coordinates:
$
    \mb{y}_a(\mb{q}) = \begin{bmatrix} \mb{0}_{4\times 1} & \mb{I}_{4\times 4} \end{bmatrix}\mb{q}.
$
The nominal controller for this system is then given by the proportional-derivative controller $\mb{k}_d(\x) = \mb{k}_{PD}(\x) \triangleq -\mb{K}_P \y(\q) - \mb{K}_D \dot{\y}(\q)$ with proportional gain $\mb{K}_P\in\mathbb{S}_{++}^4$ and derivative gain $\mb{K}_D\in\mathbb{S}_{++}^4$. 



\subsection{Control Barrier Functions for Stepping-Stones}

\noindent The stepping-stone problem is captured through the use of \textit{virtual} stepping-stones, which shrink over the course of a step to confine foot placement to a safe region defined on a targeted stone \citep{grandia2020multi}. The CBFs used to specify these foot position constraints are given by:
\begin{align}\label{eqn:ss-CBF1}
    h_1(\q) &= R(\tau(\q))-(O_x-F_x(\q)),\\
    h_2(\q) &= R(\tau(\q))+(O_x-F_x(\q)),\label{eqn:ss-CBF2}
\end{align}
where $F_x(\q)$ is the horizontal position of the swing foot and $O_x>0$ is the horizontal position of the center of stepping-stone. The virtual stone width is given by the function $R:\R\to \R$:
\begin{equation}
    R(\tau(\q)) = \frac{ar-1}{1  +ar(e^{-m(\tau(\q)-1)}-1)}+1+r
\end{equation}
where  $m>0$ determines the decay rate of the barrier function, $(1+a)r$ is half of the targeted stone width, and $1+r$ defines the half the width of the virtual stepping-stone when $\tau = 0$. These functions are visualized in Figure \ref{fig:amber_and_barrier}.
The safety constraints can be interpreted as keeping the swing foot horizontal position in an interval centered at the middle of the stepping-stone, where the interval shrinks as $\tau$ increases.
%
%
%
%
As this formulation of CBFs is position-based and therefore relative degree two, we employ the exponential control barrier function (ECBF) extension technique \citep{nguyen2016exponential} to both CBFs to attain the relative degree 1 CBFs: $h_{e,i}(\x) \triangleq L_{\f}h_i(\x)  + \alpha_{e}h_i(\q).$ 

Combining the results in Section \ref{sec:learning} and \ref{sec:bipeds}, with $\hat{\f}$ and $\hat{\g}$ the nominal model in \eqref{eq:hybrid_model_1}, the final Stepping Stone QP \eqref{SS-QP} controller combines the robustifying term of \eqref{eqn:delta-CBF-QP} with the stepping-stone ECBF extensions of \eqref{eqn:ss-CBF1} and \eqref{eqn:ss-CBF2}: 
\begin{align}
    \mb{k}(\mb{x}) =  \,\,\underset{\mb{u} \in \R^m}{\argmin}  &  \quad \frac{1}{2} \| \mb{u}-\mb{k}_{PD}(\mb{x}) \|_2^2 \tag{SS-QP} \label{SS-QP}\\
    \mathrm{s.t.} \quad & \quad L_{\hat{\f}}^2h_1(\x)+L_{\hat{\g}}L_{\hat{\f}}h_1(\x)\u + \alpha_e L_{\hat{\f}}h_1(\x) + \widehat{\delta}_1(\x) 
    \geq -\alpha(h_{e,1}(\x)) \label{finalCBF1}\\
    & \quad L_{\hat{\f}}^2h_2(\x)+L_{\hat{\g}}L_{\hat{\f}}h_2(\x)\u + \alpha_e L_{\hat{\f}}h_2(\x) + \widehat{\delta}_2(\x) 
    \geq -\alpha(h_{e,2}(\x)). \label{finalCBF2}
\end{align}



\noindent We assume that this \eqref{SS-QP} is feasible and we encountered no infeasibilities in simulation or experimentation. 
\section{Simulation and Experimental Validation} 
\label{sec:exp} 
In this section we apply our episodic learning framework (Algorithm \ref{alg:episodic_learning}) to the AMBER-3M platform in both simulation with injected model uncertainty and on hardware with the model error inherent to real-world systems. In each instance the estimator $\widehat{\delta}$ was implemented as a neural network with two hidden layers of 50 hidden units using the ReLU activation function. The network was trained minimizing mean absolute error using mini-batch gradient descent. Mean absolute error was chosen over other loss functions for its robustness to outliers. The same controller (\ref{SS-QP}) was deployed in the RaiSim \citep{raisim} simulation environment and on the AMBER-3M hardware platform, as seen in the supplementary video (\cite{videoVideo}).
The complete learning code used in simulation and experiments can be found at \citep{git}.
\begin{figure}[b!]
\centering
\includegraphics[width=\textwidth]{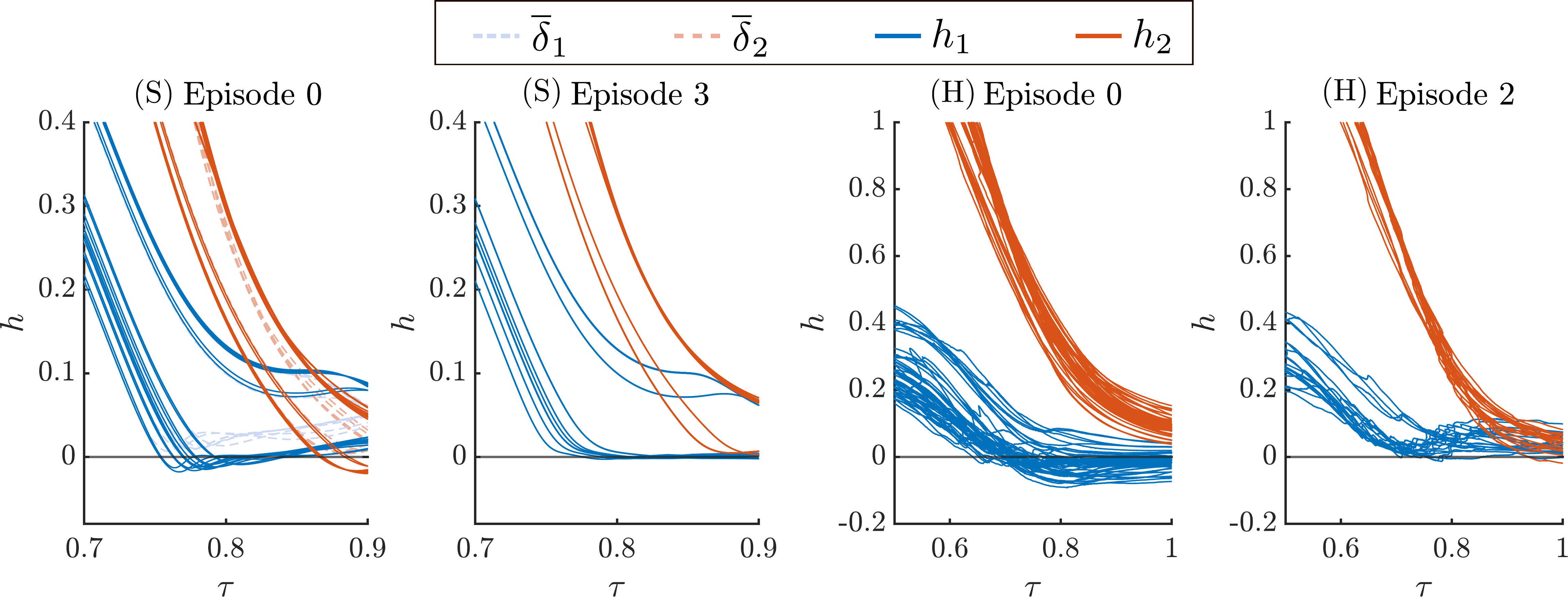}
\caption{Simulation (S) and Hardware (H) data where model mismatch causes violations. \textbf{(Far-Left)}: Simulation where the barrier functions $h_1$ (solid blue) and $h_2$ (solid orange) are enforced via a \ref{eqn:CBF-QP}. The \ref{eqn:dumbreject} is also shown for $\overline{\delta}_1$ (dashed blue) and $\overline{\delta}_2$ (dashed orange), which results in more conservative behavior over many steps. \textbf{(Mid-Left)}: After three episodes of learning the \ref{SS-QP} in simulation, the maximum barrier violation decreases from 2.0 to 0.3 cm. \textbf{(Mid-Right)}: Hardware where the barrier functions $h_1$ (blue) and $h_2$ (orange) enforced via a \ref{eqn:CBF-QP}. \textbf{(Far-Right)}: After two episodes of learning on hardware, the maximum barrier violation decreases from 9.2 to 1.9 cm via the \ref{SS-QP}.}
\label{fig:eps}
\end{figure}

\subsection{Simulation}
The controllers and learning algorithm were first validated in simulation. Model error was introduced by increasing the inertia of all limbs on the true model by a factor of ten while maintaining constant mass. Due to the underactuated nature of the robot and the relationship between step length and zero dynamics stability, not every set of stepping stones is navigable, even if safety is perfectly enforced with respect to the CBFs. Therefore, a feasible stepping stone configuration was first generated for the robot to traverse with stones of 4 cm in width. Without knowledge of the modified model ($\widehat{\delta}_1(\x) = \widehat{\delta}_2(\x) = 0$), the controller did not satisfy the CBF constraints (\ref{finalCBF1}-\ref{finalCBF2}), resulting in a maximum violation at foot placement of 2.0 cm, causing the robot to miss the stepping stone and fall over. Three episodes of the PDL algorithm were run, after which the maximum violation was reduced to be 0.3 cm, only 15\% of the original violation. Additionally, the \ref{eqn:dumbreject} controller was implemented, which ensured safety but resulted in extremely conservative behavior, resulting in poor qualitative walking, i.e. harsh foot strikes and an over-bending torso. A comparison of the barrier functions $h_1$ and $h_2$ over the steps with these controllers can be seen in Figure \ref{fig:eps}. 
\subsection{Hardware}
\begin{figure}[t!]
    \centering
    \includegraphics[width=\textwidth]{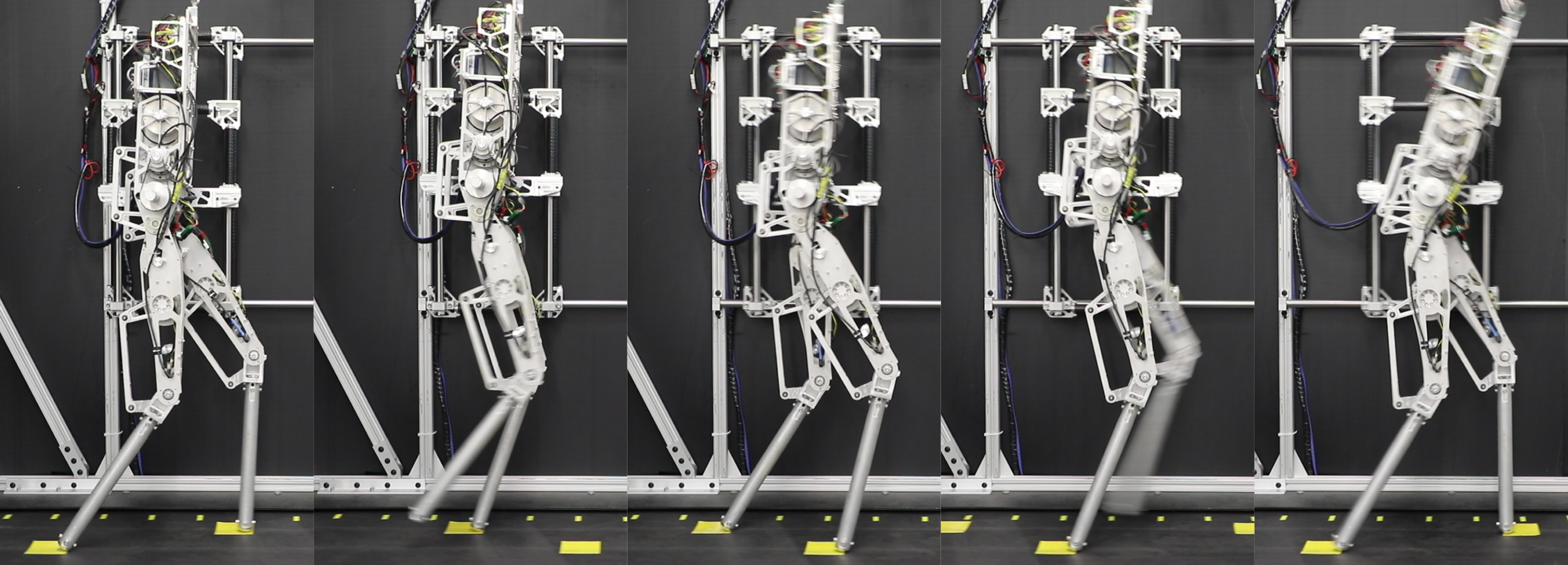}
    \caption{Gait tiles for Episode 2 of learning showing the AMBER-3M robot safely traversing a set of stepping stones. Notice the change in step width and added lean of the torso induced by the barrier functions.}
    \label{fig:gaitTiles}
    \vspace{-7mm}
\end{figure}

The same nominal model for the robot was used in the hardware experiments as in simulation, with model uncertainty presenting itself as significant friction in the joints, as well as imperfect mass and inertia measurements. The PDL algorithm was implemented on the AMBER-3M robot across a sequence of two episodes. The controllers ran on an off-board i7-6700HQ CPU @ 2.6GHz with 16 GB RAM, which computed desired torques and communicated them with the ELMO motor drivers on the 137 cm tall, 22 kg robot. The motor driver communication ran at 2kHz, and the \ref{SS-QP} ran at 1kHz.
The stepping stone configuration was specified to the controller with stones of 8 cm in width. 
As with simulation, the CBF-QP resulted in a maximum violation of the barriers of 9.2 cm due to model error. After running the PDL algorithm for two episodes, the maximum violation of the barriers was 1.9 cm, only 21\% of the original violation, as depicted in Figure \ref{fig:eps}. Although learning improves our estimate of the safe set size, as discussed in Section \ref{sec:learning}, there is still uncertainty in the stone size. This was accommodated by utilizing physical stones which were 10 cm in width.
%
%
The 7.3 cm reduction in stone size mismatch captures the change from $|\delta|_\infty$ to $|\delta-\widehat{\delta}|_\infty$. Gait tiles for this improved traversal of the stepping stones are shown in Figure \ref{fig:gaitTiles}.

\section{Conclusion}
\label{sec:conclude}
In this paper we presented an episodic learning approach for reducing the impact of model uncertainty on safety-critical control using control barrier functions. Our method is able to learn a projected disturbance and incorporate the learned model information into an optimization-based controller, as demonstrated in both a high-fidelity simulation and on the AMBER-3M robot hardware platform. Future work includes theoretical analysis of the convergence properties of Algorithm \ref{alg:episodic_learning}, modification of the controller to ensure trajectories remain close to previously-seen data points to reduce the effects of generalization error, and an extension to three dimensional bipedal walking with vision sensing for stepping-stone identification.

%
%
\clearpage
\bibliography{taylor_main}



\end{document}